\documentclass[11pt]{article}

\usepackage[final]{acl}

\usepackage{times}
\usepackage{latexsym}

\usepackage[T1]{fontenc}

\usepackage[utf8]{inputenc}

\usepackage{microtype}

\usepackage{inconsolata}

\usepackage{graphicx}

\usepackage{tabularray}
\usepackage{tabularx}
\usepackage{subcaption}
\usepackage{enumitem}
\usepackage{booktabs}
\usepackage{multirow}
\usepackage[table,xcdraw]{xcolor}
\usepackage[normalem]{ulem}
\useunder{\uline}{\ul}{}

\title{FigSIM: A Dataset for Fine-grained Suicide Severity and Figurative Language in Suicide Memes\thanks{This paper has been accepted to Findings of the Association for Computational Linguistics: ACL 2026.}}

\author{
 \textbf{Liuliu Chen\textsuperscript{1}},
 \textbf{Elise R. Carrotte\textsuperscript{2,3}},
\textbf{Brian E. Chapman\textsuperscript{4}},
 \textbf{Jo Robinson\textsuperscript{2,3}},
 \textbf{Mike Conway\textsuperscript{1}},
\\
\\
 \textsuperscript{1}School of Computing and Information Systems, University of Melbourne, Australia,
 \\
\textsuperscript{2}Orygen, The National Centre of Excellence in Youth Mental Health, Australia,
\\
 \textsuperscript{3}Centre for Youth Mental Health, University of Melbourne, Australia,
 \\
 \textsuperscript{4}O'Donnell School of Public Health, UT Southwestern Medical Center, United States
\\
 \small{
\textbf{Correspondence:} \href{mailto:liuliuc@student.unimelb.edu.au}{liuliuc@student.unimelb.edu.au}
 }
}

\begin{document}
\maketitle
\begin{abstract}

Suicide memes are memes used to express suicide-related thoughts or comment on suicide-related issues. Suicide memes are increasingly common on social media, yet remain poorly understood and potentially harmful. There is an urgent need to better understand their characteristics and to develop appropriate content moderation strategies that limits users' exposure to potentially harmful content. Currently, the absence of annotated datasets of suicide memes remains a key barrier to developing and evaluating automated moderation approaches.
In this paper, we introduce FigSIM, the first dataset designed for fine-grained analysis of suicide memes. The dataset consists of 1049 memes, each annotated for (1) fine-grained suicide severity levels, (2) figurative phenomena (e.g. metaphors), and (3) suicide-related content (e.g. suicide method depiction). We benchmark 16 unimodal and multimodal models across three tasks: figurative language, suicide severity, and suicide-related content detection. Overall, FigSIM demonstrates that suicide memes pose unique challenges for both modeling and content moderation.  Analysis revealed biases, such as underprediction of higher suicide severity levels, especially for figurative memes. The dataset (including splits used for analyses) is available at: \href{https://github.com/LiuliuChen/FigSIM}{https://github.com/LiuliuChen/FigSIM}.


\textcolor{red}{\textbf{Content Warning}: This paper contains suicide-related content that may be triggering.}

\end{abstract}

\section{Introduction}
Memes, as a multimodal form of online expression, are increasingly attracting attention from social media researchers  \cite{nguyen2024computational}. By combining visual templates with user-generated text, memes are widely used to express ideas, emotions, and humor across diverse online communities and topics \cite{literat2019youth}. However, their rapid dissemination also raises risks, especially when they are used to spread harmful content, whether deliberately (e.g., toxic, racist) or unintentionally through engagement with sensitive topics. In particular, mental health memes, especially those involving suicide-related content, present complex challenges for both research and social media moderation. 

Suicide is a serious public health problem with cascading impacts on families, communities, and society \cite{pirkis_preventing_2024}, and 
has become the third leading cause of death among 15–29-year-olds worldwide \cite{whoSuicide}. Social media plays an increasingly important role in suicide prevention \cite{rice_online_2016}. 
Young people use social media to seek help for mental health concerns, disclose suicidal thoughts, and connect with peers \cite{sala_online_2025}. These behaviors offers opportunities to better understand suicidal behavior and identify risk signals for early intervention \cite{rice_online_2016}. However, research consistently shows that exposure to suicide-related content online could be distressing or harmful for some audiences, especially when posts include explicit, graphic, or method-related details \cite{robinson_chatsafe_2018}. Such content can be emotionally challenging and triggering \cite{susi_research_2023}, which increase the likelihood of imitative behavior (e.g., self-harm) \cite{robinson_how_2025, susi_research_2023}. Importantly, the impact of suicide-related content may differ between individuals, and content shared without harmful intent may still inadvertently cause distress to those who are exposed to it \cite{thorn_motivations_2023}. 

The use of suicide memes further blurs the boundary between harmful and supportive communication. The inherent humor characteristic of memes makes it challenging to interpret the images' underlying intent \cite{vasquez_cats_2021}. Suicide jokes are often perceived as dark humor or hyperbole that may not reflect actual suicidality \cite{smith_suicide-memes_2021}. Further, meme templates are typically visually stylized \cite{nguyen2024computational}, and suicide-related content (e.g., methods) may be depicted in stylized forms that affect interpretation. Prior studies report mixed impact of suicide memes: while they may serve as a coping mechanism, they can also obscure personal distress, decrease sensitivity to suicide, or cause negative emotions among viewers \cite{perez_suicide_2019, nicomedes_convergent-mixed_2024}. This ambiguity introduces additional challenges for accurate moderation, for both human reviewers and automated systems. Yet, there is no moderation strategy or algorithm that is specifically targeted at suicide memes.

Over the last decade, studies on suicide risk on social media have focused primarily on text \cite{ji_suicidal_2021, matero_suicide_2019, sawhney_time-aware_2020}, and more recently on multimodal data such as naturalistic images and audio \cite{chatterjee_suicide_2022, badian_social_2023}, typically as binary classification tasks (suicidal vs. non-suicidal). However, assessing different levels of suicide severity risk is crucial for designing more precise and effective interventions \cite{gaur_knowledge-aware_2019}. The meme format further increases the difficulty of accurately interpreting such content, as traditional machine learning models often struggle to capture implicit cues such as metaphor and irony \cite{mazhar2025figurative}. Within the mental health domain, prior research has explored the detection of depressive and anxiety symptoms in memes \cite{huang2024towards, mazhar2025figurative}. Recent quantitative studies have revealed the distinguishable characteristics of suicide memes \cite{chen2025}. More broadly, numerous studies have focused on the detection of hateful \cite{huang2024towards}, toxic memes \cite{kiela2020hateful}, and figurative detection of political memes \cite{liu2022figmemes}.


However, to the best of our knowledge, no existing studies have yet focused on (1) identifying figurative language in suicide memes, (2) identifying fine-grained suicide risks within suicide memes, (3) assessing the transferability of current memes classification models to suicide meme. 

To address the above gaps, we present the first novel and challenging multi-label suicide memes dataset called \textbf{FigSIM} (\textbf{Fi}ne-\textbf{g}rained \textbf{Fig}urative \textbf{S}u\textbf{i}cide \textbf{M}eme). We propose three tasks: figurative language, suicide severity risk, and suicide-related content detection in suicide memes. We benchmark \textbf{FigSIM} against 16 competitive baseline models (20 configurations). 
The best-performing multimodal model achieves macro-F1 scores of 70.21\%, 71.60\%, and 58.51\% on the three tasks, respectively. Further analysis highlights the difficulty and challenges posed by suicide memes. In particular, we make the following \textbf{key contributions:} \\
\textbf{(1)} A novel dataset consists of 1,049 suicide memes with fine-grained manual annotations related to figurative phenomena, suicide severity grounded in clinical scales, and suicide-related content. \\ 
\textbf{(2)} A comprehensive benchmark of unimodal and multimodal models on proposed tasks.\\
\textbf{(3)} Thorough analyses of the dataset and benchmark results, highlighting the challenges and biases of current models on suicide memes.

\section{Related Work}

\subsection{Meme Understanding}
Memes have become widely used for communication on social media, especially among young people \cite{ypulse_2019}. Existing studies on memes can be broadly divided into: classification, interpretation, and explanation \cite{nguyen2024computational}. In this work, we focus specifically on meme classification.
Classification studies have mainly focused on detecting harmful memes, such as those that are abusive \cite{das2023banglaabusememe, jha_meme-ingful_2024}, hateful \cite{kiela2020hateful}, or offensive \cite{suryawanshi2020multimodal}, in the context of binary classification. Other works have also explored additional aspects of memes through multi-class classification, such as figurative language types (e.g., irony, metaphor) \cite{liu2022figmemes}, mental health symptoms \cite{yadav_towards_2023, mazhar2025figurative}, humor \cite{tanaka2022learning}, and targets (e.g., race) \cite{mathias2021findings}. 


\subsection{Mental health and Social Media}
There is extensive research on mental health in social media contexts. Many studies have analyzed language patterns and online engagement to identify symptoms of depression \cite{de2013predicting}, anxiety \cite{shen2017detecting}, and suicide ideation \cite{matero_suicide_2019, ji_suicidal_2021}. A few efforts have been made to classify fine-grained depressive symptoms using post text \cite{yadav2020identifying} and suicide risk severity \cite{gaur_knowledge-aware_2019}. Beyond text, researchers have also explored using visual information in shared images to capture psychological cues \cite{reece2017instagram, manikonda2017modeling}. 
Recent work has increasingly adopted multimodal approaches combining language, vision, and speech to improve detection performance \cite{chatterjee_suicide_2022, badian_social_2023}.

However, memes as an emerging multimodal social media format have received limited attention in mental health research. Only a few recent studies have developed models that use mental health memes to identify fine-grained depressive or anxious symptoms \cite{yadav_towards_2023, mazhar2025figurative}. \citet{chen2025} examined the unique characteristics of suicide-related memes and developed a binary classification model. Yet, the literature still lacks a deeper understanding of suicide meme and robust fine-grained classification models designed specifically for them.



\subsection{Suicide Memes Moderation}



Beyond detection, designing effective moderation strategies for suicide memes remains an open problem. Existing guidelines for online suicide communication, such as \#chatsafe \cite{robinson2023chatsafe}, rarely address meme-based content, largely due to the lack of empirical evidence regarding which types of suicide memes should be considered harmful. Coarse binary classification provides insufficient granularity to understand the diverse intent and impacts of suicide-related memes \cite{chen2025}. In addition, the mixed impacts of suicide memes further complicate moderation decisions \cite{mueller_suicidal_2015}.

Therefore, there is a clear gap in existing moderation models and frameworks for addressing suicide memes. An important goal for researchers, social media platforms, and society is to design and develop a context-aware and domain-specific moderation framework that can shape safer online environments. As a first step, we introduce the \textbf{FigSIM} dataset, a fine-grained multimodal dataset that captures figurative phenomena and suicide risks in memes. We hope this dataset will provide a foundation for future research on multimodal understanding and fine-grained suicide risk detection, and provide insights for designing moderation frameworks specifically targeting suicide memes.

\begin{figure*}[t]
    \centering
    \includegraphics[width=.9\linewidth]{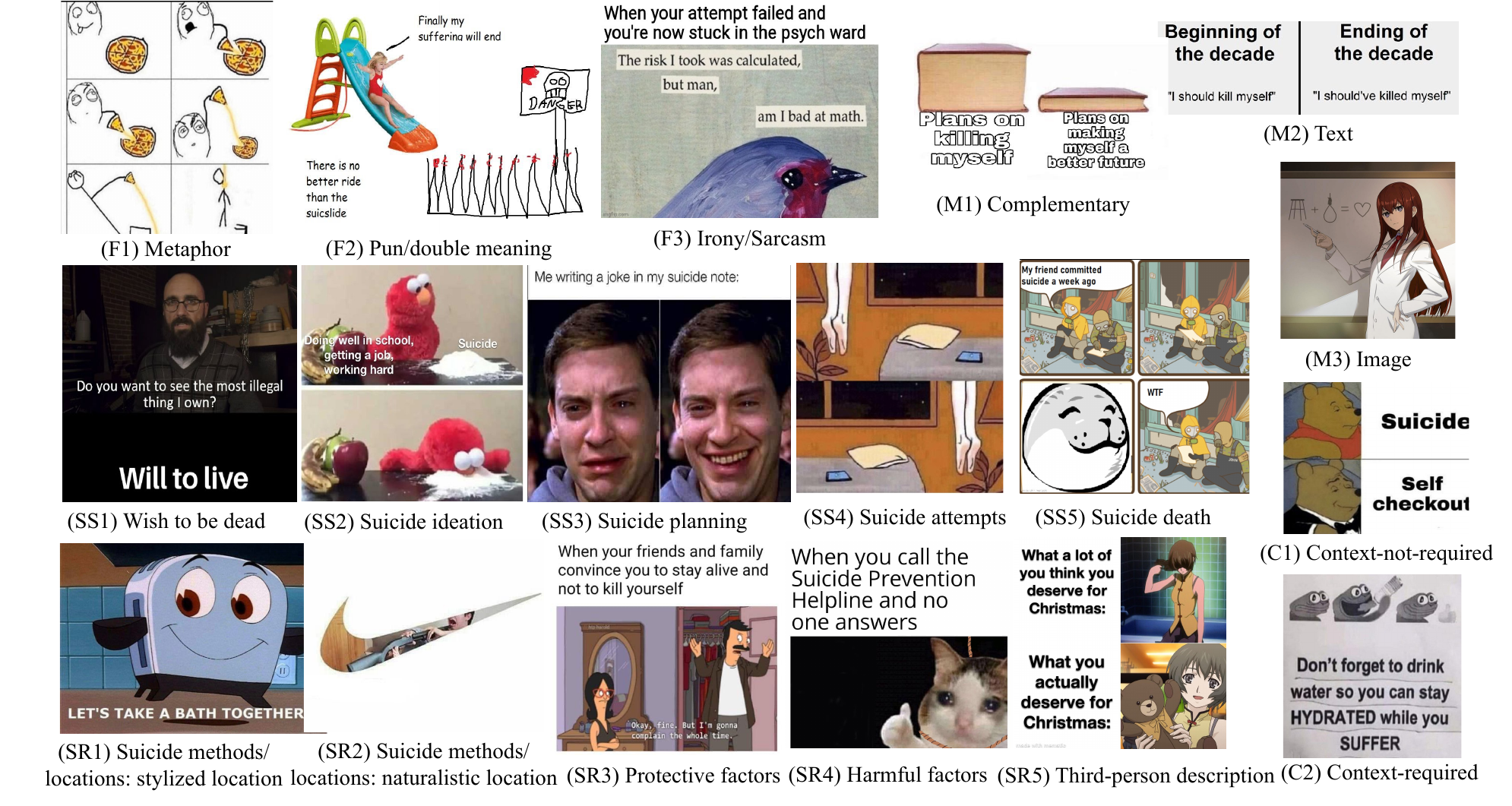}
    \caption{Examples of memes from each annotation category: Figurative Phenomenon (F1–F3), Suicide Severity (SS1–SS5), Suicide-related Content (SR1–SR5), Modality (M1–M3), and Context (C1–C2).}
    \label{fig:meme_cat_exp}
\end{figure*}

\section{Dataset Construction}
In this section, we present a new benchmark dataset: \textbf{FigSIM}, for identifying \textbf{Fi}ne-\textbf{g}rained \textbf{Fig}urative \textbf{S}u\textbf{i}cide \textbf{M}emes. The dataset was developed based on clinical assessment scales, online suicide communication guidelines, and figurative language theories. Psychologists and researchers specializing in youth suicide prevention contributed to each phase of the annotation development and process.

\subsection{Dataset Collection}
We collected images from \texttt{r/SuicideMeme}, a subreddit dedicated to sharing suicide-related memes that ranks among the top 5\% of Reddit communities by size \cite{SuicideMeme}. Using Pushshift Reddit API \cite{baumgartner2020pushshift}, we retrieved all submissions, extracted image URLs, and downloaded images using the Requests library. We filtered memes by requiring embedded text (if any) to be in English and readable by OCR, and by removing near-duplicate memes (same image with fewer than three-word differences). Memes without overlay text were kept, as purely visual memes are also important and often more difficult to interpret. After filtering, we obtained 1967 suicide memes posted between April 2018 and December 2022. Due to limited annotation resources, we randomly sampled 1050 memes for annotation.

\subsection{Annotation Scheme and Categories}
We developed a multidimensional annotation scheme consisting of the following \textbf{five categories}: \textbf{(1) figurative phenomenon}, \textbf{(2) suicide severity scale}, \textbf{(3) suicide-related content}, \textbf{(4) modality}, and \textbf{(5) context}. Figure \ref{fig:meme_cat_exp} shows the examples for each label in each category. The detailed label definitions for each category see Appendix \ref{cat_def}.

\textbf{Category: Figurative Phenomenon. }
This category captures whether a meme contains figurative expressions. The labels include: (1) \textit{Metaphor}; (2) \textit{Pun/Double meaning}; (3) \textit{Irony/Sarcasm}. Memes are labeled as \textit{None} if no figurative expressions apply. As a meme may contain multiple figurative types, this category is treated as multi-label.

\textbf{Category: Suicide Severity Scale. }
The suicide severity scale category was designed using the Columbia-Suicide Severity Rating Scale (C-SSRS) \cite{posner2011columbia}, a clinical scale widely used to assess the presence and intensity of suicidal ideation and behavior. An additional label \textit{suicide death} is introduced to better capture content referencing fatal outcomes. The final labels include: (1) \textit{None}; (2) \textit{Wish to be dead}; (3) \textit{Suicide ideation}; (4) \textit{Suicide planning}; (5) \textit{Suicide attempts}; (6) \textit{Suicide death}. This category is treated as ordinal.

\textbf{Category: Suicide-related Content. }
To enhance the scheme's applicability and capture the nature of memes and social media platforms, we designed another category: \textbf{suicide-related content}, which is treated as multi-label. We used two media guidelines: \#chatsafe \cite{robinson2023chatsafe} and the Mindframe \cite{mcternan2018media}, which were developed to support safe online communication about suicide. Additional labels were introduced to capture potential cues that are harmful or protective in the context of suicide. The final labels include: (1) \textit{Suicide methods/locations: stylized depiction}; (2) \textit{Suicide methods/locations: naturalistic depiction}; (3) \textit{Protective factors}; (4) \textit{Harmful factors}; (5) \textit{Third-person description}. Memes are labeled as \textit{None} if none apply.

\textbf{Category: Modality and Context. }
Two categories, Modality and Context, were introduced for computational analysis purposes.
For \textbf{Modality}, memes were labeled as \emph{text}, \emph{image}, or \emph{complementary}, depending on whether the meme can be interpreted using a single modality or requires both image and text jointly.
The \textbf{Context} category identifies whether interpreting suicide-related information is straightforward or requires external or background knowledge (e.g., understanding a suicide-related metaphor). Each meme is labeled as either \emph{context-required} or \emph{no-context-required}.

\subsection{Annotation Process}
We used Argilla\footnote{\href{https://argilla.io/}{https://argilla.io/}} as our annotation platform, hosted in the secure cloud environment at the University of Melbourne. 

\textbf{Pilot Annotation and Scheme Refinement.}  We conducted two pilot rounds (50 memes each) to refine the annotation scheme. Four annotators, a registered psychologist and researcher specializing in youth suicide prevention, and three computing and health informatics researchers, independently annotated the pilot batches. We held regular adjudication meetings to resolve disagreements, ensure conceptual clarity, and align interpretations, which informed iterative scheme refinement. During the pilot rounds, we documented brief image descriptions and annotation explanations to capture annotators’ reasoning (Appendix \ref{sec:meme_explanation}). Prior to large-scale annotation, a calibration batch (N=50) was annotated to assess inter-annotator agreement (IAA) between annotators.

\textbf{Large-Scale Annotation and Adjudication.} 
For large-scale annotation, two annotators from the pilot rounds (one from psychology, one from computing) independently labeled each meme. Disagreements were adjudicated by a third annotator from the pilot study, who determined the final gold-standard labels. IAA was monitored for each batch to track annotation consistency.

\setlength{\tabcolsep}{1mm}
\begin{table}[!t]
\centering
\fontsize{9pt}{9pt}\selectfont
\begin{tabular}{p{0.4\linewidth} l c p{0.25\linewidth}}
\toprule
\textbf{Category} & \textbf{Type} & \textbf{$\kappa$} & \textbf{Agreement} \\
\midrule
Suicide Severity Scale            & Single & 0.65 & Substantial \\
Modality                 & Single & 0.88 & Almost perfect \\
Context                  & Single & 0.58 & Moderate \\
Figurative Phenomenon     & Multi & 0.56 & Moderate \\
Suicide-Related Content  & Multi & 0.63 & Substantial \\
\bottomrule
\end{tabular}
\caption{IAA across annotation dimensions. Interpretation following \citet{landis1977measurement}.}
\label{tab:iaa}
\end{table}

\textbf{IAA.}
We computed IAA separately for each annotation category. For single-label categories, we used Cohen’s $\kappa$ \cite{cohen1960coefficient} to measure chance-corrected agreement. For multi-label categories, we report the average Cohen’s $\kappa$ across individual labels (excluding \emph{None}). Final agreement was calculated on the independently annotated large-scale dataset (N=900), excluding pilot and calibration batches, as shown in Table~\ref{tab:iaa}.

\section{Dataset Analysis}
The final dataset contains 1049 suicide memes annotated across five categories: figurative phenomenon, suicide severity, suicide-related content, modality, and context. One meme was discarded due to quality control. In this section, we analyze the dataset to highlight its complexity and associated modeling challenges.

\subsection{Data Distribution and Statistics}
\begin{figure*}[!tp]
    \centering
    \includegraphics[width=\linewidth]{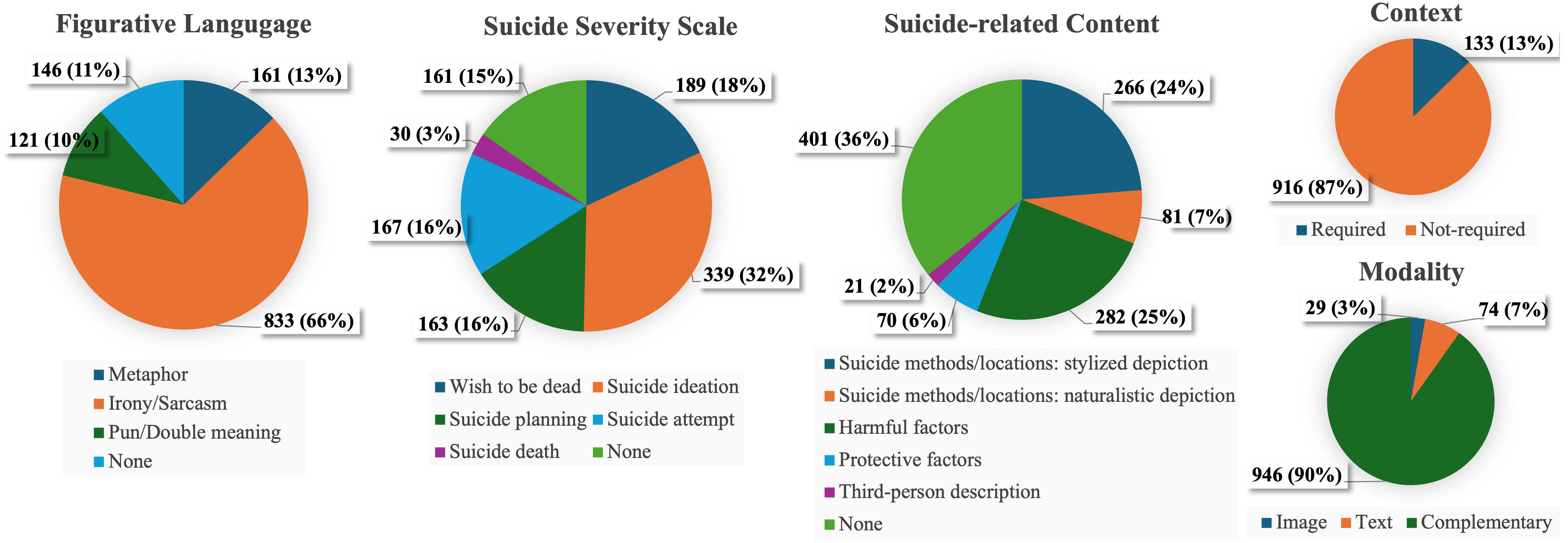}
    \caption{Distribution of annotations across all categories in the \textbf{FigSIM} dataset. 
Pie charts show the label frequencies and proportions for each annotation category.
}
    \label{fig:dataset_stat}
\end{figure*}

Figure \ref{fig:dataset_stat} shows the label distributions across annotation categories. For Suicide Severity Scale, the dataset is dominated by \emph{suicide ideation}, while very few memes indicate \emph{suicide death}. Most memes are labeled as \emph{irony/sarcasm}, while 146 out of 1049 contain no figurative expression. For Suicide-related Content, \textit{stylized method depictions} and \textit{harmful factors} are most common, whereas \textit{protective factors} and \textit{third-person descriptions} occur less often. On average, each meme contains 1.06 figurative labels and 0.69 suicide-related labels (excluding \emph{None}). Most memes are \emph{complementary} and \emph{no-context-required}. Cross-category relationships are analyzed in Appendix \ref{sec: app_data_stat}.

\subsection{Inter-annotator Agreement}
Table~\ref{tab:iaa} reports IAA across the annotation categories. Overall, Suicide Severity Scale, Suicide-related Content, and Modality show substantial to almost perfect agreement, suggesting consistent agreement in identifying suicide-related risks and the visual–textual structure of memes. In contrast, Figurative Phenomenon and Context show lower agreement (moderate), potentially due to the inherent subjectivity of figurative interpretation and the nuanced judgments required to determine whether external context is needed, which may also vary across annotator demographics. Supplementary metrics and label-wise agreement see Appendix \ref{sec: iaa}.

\subsection{Visual and Language Analysis}
We conducted a visual and language analysis focusing on the category Suicide Severity Scale.

\textbf{Visual Analysis:} We extracted image features, including brightness, colors (HSV), contrast (standard deviation of grayscale values), and face presence (binary Haar-cascade detection). We compared these features across suicide severity scale labels using Kruskal–Wallis tests with Dunn’s post-hoc comparisons (Bonferroni adjusted, $\alpha=.05$). Two features showed significant differences: \emph{Suicide death} memes show lower saturation than all other labels, while contrast is significantly higher in \emph{suicide planning} and \emph{suicide death} memes compared to \emph{wish to be dead} memes. Other features also showed non-significant but interesting trends. For example, face presence was lower in \emph{suicide attempts} than in \emph{suicide ideation} ($p=.063$).

\textbf{Language Analysis:} Using LIWC \cite{pennebaker2015development} and EmoLex \cite{Mohammad13_nrc} lexicons, we quantified psycholinguistic and emotional features on meme text. The same statistical tests used in the visual analysis were applied. Overall, 35 features differ significantly across the Suicide Severity Scale. Linguistic complexity varies across severity. For instance, \emph{suicide planning} memes contain longer sentences than \emph{wish to be dead} and \emph{suicide attempts}, while \emph{suicide ideation} shows higher word count than \emph{suicide attempts}. Affective features, such as sadness 
and fear, are more prevalent in \emph{suicide ideation} memes than in other levels. Finally, death-related terms show the strongest distinctions across 
severity levels, with higher usage in \textit{suicide ideation} memes.

\subsection{Moderation Analysis}
For moderation analysis, we evaluate three multimodal moderation models: \texttt{OpenAI omni-moderation-latest}, \texttt{Azure AI Content Safety}, \texttt{Llama-Guard-3-8B}. Although our dataset lacks harmfulness labels, these models’ predictions provides insights to how current moderation models behave on suicide memes. As shown in Table \ref{tab:mod_res}, the proportion of flagged memes increases with suicide severity across all models. Azure AI Content Safety is the most sensitive, with the highest flag rates for both suicide/self-harm and overall harm categories. Notably, the overall harm flag rate is consistently higher than the suicide/self-harm flag rate, indicating that some memes are flagged due to other potential risks (e.g., violence) rather than suicide-related concerns.

\begin{figure}[t]
    \centering
    \includegraphics[width=\linewidth]{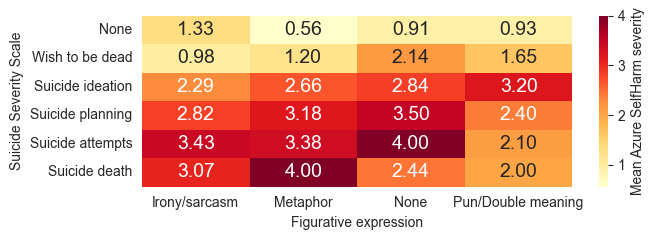}
    \caption{Azure moderation outcomes by suicide severity and figurative expression. Values represent mean self-harm severity score (scale: 0-7).}
    \label{fig:azure_heatmap}
\end{figure}

Figure \ref{fig:azure_heatmap} shows Azure's suicide/self-harm severity score across suicide severity levels and figurative phenomenon. At the same severity level, figurative memes tend to receive lower scores than non-figurative memes. The results suggest that figurative language may impact how moderation models interpret suicide-related content.

\setlength{\tabcolsep}{1mm}
\begin{table}[t]
\centering
{\fontsize{9pt}{9pt}\selectfont
\begin{tabularx}{\linewidth}{@{}l X X l X X X@{}}
\toprule
 & \multicolumn{3}{c}{\textbf{Suicide/Self-harm}} 
 & \multicolumn{3}{c}{\textbf{Overall Harm}} \\
\cmidrule(lr){2-4} \cmidrule(lr){5-7}
 & Azure & Llama & OpenAI
 & Azure & Llama & OpenAI \\
\midrule

None &
  \cellcolor[HTML]{FFFFC7}0.304 &
  \cellcolor[HTML]{FFFFC7}0.304 &
  \cellcolor[HTML]{FFFFC7}0.143 &
  \cellcolor[HTML]{ECF4FF}0.437 &
  \cellcolor[HTML]{ECF4FF}0.304 &
  \cellcolor[HTML]{ECF4FF}0.180 \\

Wish to be dead &
  \cellcolor[HTML]{FFFFC7}0.271 &
  \cellcolor[HTML]{FFFFC7}0.312 &
  \cellcolor[HTML]{FFFFC7}0.243 &
  \cellcolor[HTML]{ECF4FF}0.372 &
  \cellcolor[HTML]{ECF4FF}0.312 &
  \cellcolor[HTML]{ECF4FF}0.339 \\

Suicide ideation &
  \cellcolor[HTML]{FFFFC7}0.619 &
  \cellcolor[HTML]{FFFFC7}0.386 &
  \cellcolor[HTML]{FFFFC7}0.475 &
  \cellcolor[HTML]{ECF4FF}0.678 &
  \cellcolor[HTML]{ECF4FF}0.386 &
  \cellcolor[HTML]{ECF4FF}0.501 \\

Suicide planning &
  \cellcolor[HTML]{FFFFC7}0.560 &
  \cellcolor[HTML]{FFFFC7}0.509 &
  \cellcolor[HTML]{FFFFC7}0.497 &
  \cellcolor[HTML]{ECF4FF}0.660 &
  \cellcolor[HTML]{ECF4FF}0.509 &
  \cellcolor[HTML]{ECF4FF}0.521 \\

Suicide attempts &
  \cellcolor[HTML]{FFFFC7}0.661 &
  \cellcolor[HTML]{FFFFC7}0.428 &
  \cellcolor[HTML]{FFFFC7}0.575 &
  \cellcolor[HTML]{ECF4FF}0.745 &
  \cellcolor[HTML]{ECF4FF}0.428 &
  \cellcolor[HTML]{ECF4FF}0.653 \\

Suicide death &
  \cellcolor[HTML]{FFFFC7}0.633 &
  \cellcolor[HTML]{FFFFC7}0.333 &
  \cellcolor[HTML]{FFFFC7}0.500 &
  \cellcolor[HTML]{ECF4FF}0.767 &
  \cellcolor[HTML]{ECF4FF}0.333 &
  \cellcolor[HTML]{ECF4FF}0.667 \\

\bottomrule
\end{tabularx}
}
\caption{Moderation flag rates across suicide severity. 
“Suicide/Self-harm” refers to flags triggered specifically by self-harm, whereas “Overall Harm” includes any flag raised by other safety categories (e.g., violence).}
\label{tab:mod_res}
\end{table}

\section{Experiments, Results, and Analysis}

\setlength{\tabcolsep}{1mm}
\begin{table*}[!t]
\centering
{\fontsize{9pt}{9pt}\selectfont
\begin{tabular}{@{}l l c c c c c c@{}}
\toprule
\multirow{2}{*}{\textbf{Input}} &
\multirow{2}{*}{\textbf{Model}} &
\multicolumn{2}{c}{\textbf{Figurative detection}} &
\multicolumn{2}{c}{\textbf{Suicide severity detection}} &
\multicolumn{2}{c}{\textbf{Suicide-related detection}} \\
\cmidrule(lr){3-4} \cmidrule(lr){5-6} \cmidrule(lr){7-8}
 &  &
F1$_{macro}$ & F1$_{weighted}$ &
F1$_{macro}$ & F1$_{weighted}$ &
F1$_{macro}$ & F1$_{weighted}$ \\

\midrule

\multirow{3}{*}{\textit{Text-only}} & BERT &
  \cellcolor[HTML]{FFFFC7}51.25$\pm$4.34 & \cellcolor[HTML]{FFFFC7}75.98$\pm$1.44 &
  \cellcolor[HTML]{ECF4FF}48.45$\pm$2.78 & \cellcolor[HTML]{ECF4FF}49.73$\pm$2.46 &
  \cellcolor[HTML]{96FFFB}35.22$\pm$3.01 & \cellcolor[HTML]{96FFFB}44.31$\pm$2.66 \\
 & MENTALBERT &
  \cellcolor[HTML]{FFFFC7}54.09$\pm$4.89 & \cellcolor[HTML]{FFFFC7}77.06$\pm$1.75 &
  \cellcolor[HTML]{ECF4FF}48.07$\pm$2.46 & \cellcolor[HTML]{ECF4FF}50.03$\pm$2.49 &
  \cellcolor[HTML]{96FFFB}40.69$\pm$1.95 & \cellcolor[HTML]{96FFFB}49.33$\pm$0.96 \\
 & ROBERTA &
  \cellcolor[HTML]{FFFFC7}53.05$\pm$1.99 & \cellcolor[HTML]{FFFFC7}76.41$\pm$0.91 &
  \cellcolor[HTML]{ECF4FF}49.97$\pm$3.39 & \cellcolor[HTML]{ECF4FF}51.81$\pm$3.20 &
  \cellcolor[HTML]{96FFFB}38.37$\pm$1.12 & \cellcolor[HTML]{96FFFB}48.58$\pm$1.01 \\

\midrule
\multirow{3}{*}{\textit{Image-only}} & ResNet-50 &
  \cellcolor[HTML]{FFFFC7}43.18$\pm$4.74 & \cellcolor[HTML]{FFFFC7}72.23$\pm$1.77 &
  \cellcolor[HTML]{ECF4FF}21.70$\pm$3.67 & \cellcolor[HTML]{ECF4FF}22.55$\pm$3.52 &
  \cellcolor[HTML]{96FFFB}25.58$\pm$1.32 & \cellcolor[HTML]{96FFFB}34.88$\pm$0.58 \\
 & ViT &
  \cellcolor[HTML]{FFFFC7}55.53$\pm$1.78 & \cellcolor[HTML]{FFFFC7}77.20$\pm$0.78 &
  \cellcolor[HTML]{ECF4FF}25.88$\pm$1.11 & \cellcolor[HTML]{ECF4FF}27.99$\pm$0.69 &
  \cellcolor[HTML]{96FFFB}32.36$\pm$3.39 & \cellcolor[HTML]{96FFFB}36.82$\pm$3.99 \\
 & DINOv2 &
  \cellcolor[HTML]{FFFFC7}53.02$\pm$2.63 & \cellcolor[HTML]{FFFFC7}76.07$\pm$1.50 &
  \cellcolor[HTML]{ECF4FF}26.75$\pm$4.75 & \cellcolor[HTML]{ECF4FF}28.13$\pm$5.73 &
  \cellcolor[HTML]{96FFFB}31.86$\pm$1.90 & \cellcolor[HTML]{96FFFB}37.09$\pm$1.62 \\

\midrule
\multirow{14}{*}{\textit{Multimodal}} & CLIP &
  \cellcolor[HTML]{FFFFC7}62.06$\pm$1.27 & \cellcolor[HTML]{FFFFC7}80.21$\pm$0.36 &
  \cellcolor[HTML]{ECF4FF}49.97$\pm$1.08 & \cellcolor[HTML]{ECF4FF}52.10$\pm$1.32 &
  \cellcolor[HTML]{96FFFB}33.70$\pm$1.67 & \cellcolor[HTML]{96FFFB}43.75$\pm$1.86 \\
 & BLIP-2 &
  \cellcolor[HTML]{FFFFC7}55.36$\pm$4.08 & \cellcolor[HTML]{FFFFC7}77.46$\pm$1.80 &
  \cellcolor[HTML]{ECF4FF}39.47$\pm$0.34 & \cellcolor[HTML]{ECF4FF}41.95$\pm$0.41 &
  \cellcolor[HTML]{96FFFB}38.81$\pm$3.37 & \cellcolor[HTML]{96FFFB}44.25$\pm$2.65 \\
 & GPT-5-mini zero-shot &
  \cellcolor[HTML]{FFFFC7}65.46$\pm$2.87 & \cellcolor[HTML]{FFFFC7}80.40$\pm$2.03 &
  \cellcolor[HTML]{ECF4FF}65.78$\pm$2.28 & \cellcolor[HTML]{ECF4FF}66.22$\pm$2.29 &
  \cellcolor[HTML]{96FFFB}52.52$\pm$3.99 & \cellcolor[HTML]{96FFFB}53.36$\pm$3.03 \\
 & GPT-5-mini few-shot &
  \cellcolor[HTML]{FFFFC7}63.58$\pm$1.46 & \cellcolor[HTML]{FFFFC7}76.04$\pm$0.91 &
  \cellcolor[HTML]{ECF4FF}67.41$\pm$1.46 & \cellcolor[HTML]{ECF4FF}67.95$\pm$1.31 &
  \cellcolor[HTML]{96FFFB}53.87$\pm$0.46 & \cellcolor[HTML]{96FFFB}54.54$\pm$0.47 \\
 & Claude-sonnet-4-5 zero-shot &
  \cellcolor[HTML]{FFFFC7}\textbf{70.21}$\pm$0.82 & \cellcolor[HTML]{FFFFC7}\textbf{80.69}$\pm$0.42 &
  \cellcolor[HTML]{ECF4FF}62.51$\pm$2.44 & \cellcolor[HTML]{ECF4FF}63.27$\pm$2.31 &
  \cellcolor[HTML]{96FFFB}55.72$\pm$1.85 & \cellcolor[HTML]{96FFFB}59.96$\pm$1.56 \\
 & Claude-sonnet-4-5 few-shot &
  \cellcolor[HTML]{FFFFC7}66.87$\pm$1.81 & \cellcolor[HTML]{FFFFC7}70.81$\pm$1.20 &
  \cellcolor[HTML]{ECF4FF}62.02$\pm$0.38 & \cellcolor[HTML]{ECF4FF}62.80$\pm$0.44 &
  \cellcolor[HTML]{96FFFB}55.15$\pm$1.50 & \cellcolor[HTML]{96FFFB}\textbf{61.21}$\pm$1.36 \\
 & Gemini-3-pro zero-shot &
  \cellcolor[HTML]{FFFFC7}60.24$\pm$0.24 & \cellcolor[HTML]{FFFFC7}73.15$\pm$0.92 &
  \cellcolor[HTML]{ECF4FF}66.09$\pm$2.21 & \cellcolor[HTML]{ECF4FF}65.67$\pm$2.24 &
  \cellcolor[HTML]{96FFFB}\textbf{58.51}$\pm$4.06 & \cellcolor[HTML]{96FFFB}58.16$\pm$2.53 \\
 & Gemini-3-pro few-shot &
  \cellcolor[HTML]{FFFFC7}63.79$\pm$1.78 & \cellcolor[HTML]{FFFFC7}74.38$\pm$1.39 &
  \cellcolor[HTML]{ECF4FF}\textbf{71.60}$\pm$2.43 & \cellcolor[HTML]{ECF4FF}\textbf{71.73}$\pm$2.52 &
  \cellcolor[HTML]{96FFFB}54.57$\pm$2.06 & \cellcolor[HTML]{96FFFB}57.03$\pm$1.26 \\
 & Llama-3.2 Vision zero-shot &
  \cellcolor[HTML]{FFFFC7}50.59$\pm$0.32 & \cellcolor[HTML]{FFFFC7}71.73$\pm$0.13 &
  \cellcolor[HTML]{ECF4FF}34.62$\pm$6.79 & \cellcolor[HTML]{ECF4FF}33.51$\pm$0.00 &
  \cellcolor[HTML]{96FFFB}23.98$\pm$3.40 & \cellcolor[HTML]{96FFFB}37.60$\pm$0.00 \\
 & Llama-3.2 Vision few-shot &
  \cellcolor[HTML]{FFFFC7}34.42$\pm$0.00 & \cellcolor[HTML]{FFFFC7}66.10$\pm$0.00 &
  \cellcolor[HTML]{ECF4FF}26.03$\pm$0.00 & \cellcolor[HTML]{ECF4FF}30.83$\pm$0.00 &
  \cellcolor[HTML]{96FFFB}22.86$\pm$0.00 & \cellcolor[HTML]{96FFFB}32.70$\pm$0.00 \\
 & QWen3-VL zero-shot &
  \cellcolor[HTML]{FFFFC7}59.13$\pm$0.11 & \cellcolor[HTML]{FFFFC7}76.48$\pm$0.14 &
  \cellcolor[HTML]{ECF4FF}42.03$\pm$0.41 & \cellcolor[HTML]{ECF4FF}44.62$\pm$0.52 &
  \cellcolor[HTML]{96FFFB}44.45$\pm$6.79 & \cellcolor[HTML]{96FFFB}49.00$\pm$0.01 \\
 & QWen3-VL few-shot &
  \cellcolor[HTML]{FFFFC7}58.85$\pm$4.33 & \cellcolor[HTML]{FFFFC7}75.85$\pm$2.95 &
  \cellcolor[HTML]{ECF4FF}52.55$\pm$0.21 & \cellcolor[HTML]{ECF4FF}54.45$\pm$0.11 &
  \cellcolor[HTML]{96FFFB}42.80$\pm$0.00 & \cellcolor[HTML]{96FFFB}48.14$\pm$0.00 \\
 & InternVL3.5 zero-shot &
  \cellcolor[HTML]{FFFFC7}42.81$\pm$4.18 & \cellcolor[HTML]{FFFFC7}74.71$\pm$5.74 &
  \cellcolor[HTML]{ECF4FF}31.03$\pm$0.00 & \cellcolor[HTML]{ECF4FF}33.87$\pm$0.00 &
  \cellcolor[HTML]{96FFFB}45.13$\pm$1.38 & \cellcolor[HTML]{96FFFB}45.91$\pm$1.78 \\
 & InternVL3.5 few-shot &
  \cellcolor[HTML]{FFFFC7}49.09$\pm$0.00 & \cellcolor[HTML]{FFFFC7}72.07$\pm$0.00 &
  \cellcolor[HTML]{ECF4FF}31.88$\pm$0.00 & \cellcolor[HTML]{ECF4FF}34.93$\pm$0.68 &
  \cellcolor[HTML]{96FFFB}45.18$\pm$1.79 & \cellcolor[HTML]{96FFFB}48.42$\pm$0.60 \\

\midrule
\multirow{2}{*}{\textit{Specialized}} & \citet{yadav_towards_2023} &
  \cellcolor[HTML]{FFFFC7}-- & \cellcolor[HTML]{FFFFC7}-- &
  \cellcolor[HTML]{ECF4FF}49.01$\pm$2.98 & \cellcolor[HTML]{ECF4FF}50.89$\pm$2.99 &
  \cellcolor[HTML]{96FFFB}-- & \cellcolor[HTML]{96FFFB}-- \\
 & M3H &
  \cellcolor[HTML]{FFFFC7}-- & \cellcolor[HTML]{FFFFC7}-- &
  \cellcolor[HTML]{ECF4FF}61.47$\pm$1.74 & \cellcolor[HTML]{ECF4FF}62.58$\pm$1.95 &
  \cellcolor[HTML]{96FFFB}-- & \cellcolor[HTML]{96FFFB}-- \\
\bottomrule
\end{tabular}
}
\caption{Comparison of model performance on the test set ($N=208$).}
\label{tab:model_result}
\end{table*}

We proposed three tasks for the \textbf{FigSIM} dataset: (1) figurative language detection, (2) suicide severity detection, and (3) suicide-related content detection. Below, we present baseline results for a set of state-of-the-art unimodal and multimodal models.

\subsection{Baselines}
We evaluate baseline models across text-only, image-only, and multimodal configurations. Our selection aims to provide a representative benchmark including widely used pretrained text and vision, multimodal models, recent multimodal large language models (MLLMs), and task-adjacent meme models from mental health research.

\textbf{Text only:} 
We extract OCR text from each meme using Google Vision API and fine-tune three pretrained language models: BERT-base \cite{devlin2019bert}, MentalBERT \cite{ji2022mentalbert}, and RoBERTa \cite{liu2019roberta}.

\textbf{Image only:} 
We fine-tune ResNet-50 \cite{he2016deep}, ViT-Base \cite{dosovitskiy2020vit}, and DINOv2 \cite{oquab2023dinov2}.

\textbf{Multimodal:} 
We finetune CLIP \cite{radford2021learning} and BLIP-2 \cite{li2023blip}. We additionally evaluate MLLMs using zero- and few-shot prompting: \texttt{gpt-5-mini} \cite{openai-2025-gpt5}, \texttt{claude-sonnet-4-5} \cite{anthropic-2025-claude-sonnet}, \texttt{gemini-3-pro} \cite{google2025gemini}, \texttt{Llama-3.2-11B-Vision-Instruct} \cite{dubey2024llama}, \texttt{Qwen3-VL-8B-Instruct} \cite{qwen-2025-qwen3-vl}, and \texttt{InternVL3\_5-8B} \cite{wang2025internvl3_5}.

\textbf{Specialized meme models:} 
We further evaluate mental-health meme models from adjacent domains to examine their cross-domain generalization to suicide memes. Specifically, we include the depression symptom detection model proposed by \citet{yadav_towards_2023} and the M3H framework \cite{mazhar2025figurative}, a recent model for mental-health meme classification that incorporates figurative and commonsense signals. 

\textbf{Implementation details}
We split the dataset into training, validation, and test sets using a stratified strategy to preserve label distributions across all annotation categories. The data was divided using a 60:20:20 ratio, resulting in 633 training memes, 208 validation memes, and 208 test memes (for detailed label-wise statistics see Appendix~\ref{sec: implementation_detail}). We trained all supervised models using CrossEntropyLoss for single-label categories and BCEWithLogitsLoss for multi-label categories. To mitigate class imbalance, we applied class-reweighted losses and tuned decision thresholds for multi-label tasks. For MLLMs, prompts were constructed using the annotation definitions and guidelines. We report macro-F1 and weighted-F1 scores on test set. For detailed training configurations see Appendix \ref{sec: implementation_detail}.

Due to extreme class imbalance, we merged \emph{Suicide Attempt} and \emph{Suicide Death} into a single class, \textbf{\emph{Suicide Attempt/Death}}, for suicide severity classification, and \textbf{exclude} the \emph{Third-person description} label from suicide-related content detection.

 \subsection{Performance Comparison}
Table \ref{tab:model_result} shows baseline model performance across tasks. Language models consistently outperform vision models on suicide severity and suicide-related content detection. This suggests that linguistic cues play an important role in inferring suicide intent, while visual information alone is often insufficient.

MLLMs achieve the strongest performance across all three tasks, with commercial MLLMs consistently outperforming open-source MLLMs. Notably, LLaMA refuses to generate predictions for several images; therefore, these instances are excluded from evaluation for LLaMA. This behavior reflects the safety filtering mechanisms of LLMs, which may limit their applicability in this context. Interestingly, few-shot prompting often reduces performance for figurative detection, likely due to prompt sensitivity and subjective interpretation. In contrast, suicide severity detection benefits more consistently from few-shot examples, which indicates that explicit guidance helps models distinguish between fine-grained risk levels. Across models, suicide-related content detection is the most challenging task, with the lowest macro-F1 scores. 

Existing models for mental-health memes detection demonstrate certain transferability for suicide severity detection. The M3H framework outperforms many baseline models, but there is still a substantial gap (10\%) between M3H and the best-performing model. This gap suggests the need for suicide-specific domain knowledge beyond what is captured by current mental-health meme models.

Table \ref{tab:per_label_f1} shows per-label F1 scores for the best-performing model selected based on macro-F1 for each task. 
For figurative detection, performance is highest for \textit{irony/sarcasm}, potentially due to the its dominance in the dataset.
Suicide severity detection shows relatively balanced performance across severity levels, with the lowest F1 in \textit{suicide planning}. 
Suicide-related content detection exhibits lower F1 scores for \textit{harmful factors} and \textit{naturalistic depictions}, indicating greater modeling difficulty.

\subsection{Error Analysis}
To better understand model performance, we conduct an error analysis on the best-performing models for each task.

\begin{figure}[!t]
    \centering
    \includegraphics[width=\linewidth]{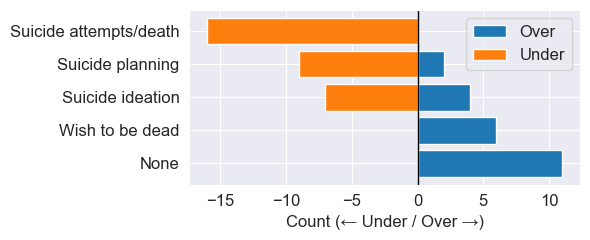}
    \caption{Counts of over- and underprediction for suicide severity, conditioned on true labels.}
    \label{fig:severity_err}
\end{figure}

\textbf{Severity calibration bias.}
Figure \ref{fig:severity_err} shows over- and underprediction of suicide severity conditioned on the true label. For suicide severity detection, models tend to over-predict lower-severity categories (\textit{None} and \textit{Wish to be dead}) but underpredict higher suicide severity (\textit{Suicide planning} and \textit{Suicide attempts/death}).
Figurative language is associated with this underprediction. When suicidal risk is expressed figuratively, severity is substantially underestimated more often than for non-figurative memes. This pattern is most evident for metaphorical expressions, though the number of such cases is limited. Importantly, this effect persists even when figurative language is correctly identified, suggesting that the primary challenge lies not only in detecting figurative cues but also in mapping indirect expressions to appropriate severity levels. Detailed analyses are provided in Appendix~\ref{sec:app_err_analysis}.

\setlength{\tabcolsep}{1mm}
\begin{table}[!t]
\centering
\small
\begin{tabular}{p{3cm} c c p{1.5cm}}
\toprule
\textbf{Label} & \textbf{F1} & \textbf{Avg. F1} & \textbf{Model} \\
\midrule
\multicolumn{4}{l}{\emph{Figurative Phenomenon}} \\
\quad Irony/sarcasm        & 86.70$\pm$0.60 & \multirow{3}{*}{70.21} & \multirow{3}{*}{\parbox[c]{1.5cm}{\raggedright\ttfamily Claude sonnet zero-shot}} \\
\quad Metaphor             & 65.93$\pm$1.28 &                      &  \\
\quad Pun/double meaning   & 58.00$\pm$1.54 &                      &  \\
\midrule
\multicolumn{4}{l}{\emph{Suicide Severity Scale}} \\
\quad Wish to be dead            & 75.68$\pm$4.80 & \multirow{4}{*}{71.60} & \multirow{4}{*}{\parbox[c]{1.5cm}{\raggedright\ttfamily Gemini 3 pro few-shot}} \\
\quad Suicide ideation           & 71.57$\pm$3.13 &                      &  \\
\quad Suicide planning           & 68.14$\pm$3.23 &                      &  \\
\quad Suicide attempt/death      & 71.62$\pm$1.92 &                      &  \\
\quad None     & 70.98$\pm$1.71 &                      &  \\
\midrule
\multicolumn{4}{l}{\emph{Suicide-related Content}} \\
\quad Protective factors                                  & 61.76$\pm$8.02 & \multirow{5}{*}{58.51} & \multirow{5}{*}{\parbox[c]{1.5cm}{\raggedright\ttfamily Gemini 3 pro zero-shot}} \\
\quad Harmful factors                                     & 47.19$\pm$1.05 &                      &  \\
\quad Stylized depiction       & 69.92$\pm$2.45 &                      &  \\
\quad Naturalistic depiction   & 55.18$\pm$5.97 &                      &  \\
\bottomrule
\end{tabular}
\caption{Per-label F1 scores and Average (per-label F1) for the best-performing model in each task.}
\label{tab:per_label_f1}
\end{table}

\textbf{Context-required cases are challenging.}
Although most misclassified memes are \textit{no-context-required} due to dataset imbalance, \textit{context-required} memes exhibit relatively higher error rates across tasks. Similarly, while complementary memes dominate the dataset, unimodal cases (e.g., text-only) show higher misclassification rates. These observations suggest that models may struggle without context or multimodal cues, though conclusions are constrained by the small number of such cases. 
Detailed analyses are provided in Appendix~\ref{sec:app_err_analysis}.

\section{Discussion and Implications}

Our dataset analysis highlights the unique challenges of suicide meme understanding. Lower IAA on figurative phenomenon and context reflects the inherent subjectivity and ambiguity of humor and implicit meaning in memes, which complicates suicide severity interpretation. Existing moderation models exhibit varying sensitivity across suicide severity levels. Memes with higher suicide severity are more likely to be flagged, while figurative expressions are often assessed as less severe. However, future qualitative validation is necessary to determine if flagged memes are truly harmful. 
Across baseline models and tasks, MLLMs achieve the strongest results, though task difficulty varies substantially within the dataset. Error analysis shows that MLLMs tend to underpredict higher severity categories, and performances may be affected when the context is required. Overall, these findings highlight the complexity of suicide memes and the need for models that better integrate figurative language, multimodal signals, and fine-grained suicide risk.

Our dataset highlights the importance of explicitly considering meme formats in online safety and suicide content moderation. We hope this dataset can support hypothesis generation and evidence-building regarding the characteristics of suicide memes, inform the development of more context-aware moderation strategies, and contribute to guidance on safer online communication about suicide beyond text-based communication.






\section{Conclusion}
We introduce \textbf{FigSIM}, the first dataset designed for fine-grained analysis of suicide memes. By integrating annotations of figurative language, suicide scale severity, and suicide-related content, FigSIM captures the complexity and ambiguity of suicide communication in meme format. Benchmark results across unimodal and multimodal models show that suicide memes pose unique challenges to existing models. Overall, FigSIM provides a foundation for studying suicide memes and for developing more context-aware detection and moderation frameworks beyond text-based communication.

\section*{Limitations}
There are several limitations in this study. First, the dataset is derived from a single data source, \texttt{r/SuicideMeme} on Reddit. Suicide-related communication, humor norms, and commonsense knowledge may vary across communities, cultures, and platforms. Future work aims to extend this dataset to additional sources to improve coverage and generalizability.

Second, while our annotation scheme was developed in consultation with youth suicide prevention researchers including a registered psychologist, the labels for figurative phenomenon and suicide severity are inherently subjective. The interpretation and annotation rely solely on the content within each meme, under the assumption that the meme reflects indirect self-expression unless explicitly indicated otherwise. Moreover, while our scheme introduces more granularity than prior work, the categories are not exhaustive and do not capture all nuances of suicide-related expression in memes. ``Fine-grained” should therefore be understood as relative rather than comprehensive.

Accordingly, the annotations should not be interpreted as clinical judgments, nor do they provide any diagnostic insights about the meme creators. Our main goal is to understand how suicidality is communicated through humor and figurative expression in online memes, to characterize their varying levels of harmfulness, and support the development of more context-aware moderation strategies.

\section*{Ethical Considerations}
This study involves the collection and analysis of suicide memes, which contain highly sensitive and potentially distressing material. Our primary motivation in constructing the dataset is to support research on the detection of suicide-related online content and to inform responsible moderation strategies. However, we acknowledge that there is potential for misuse by bad actors in ways we cannot fully anticipate.

All data used in this study were collected exclusively from the publicly accessible subreddit \texttt{r/SuicideMemes}. One ethical challenge is that we can not seek individual consent from Reddit users. To better protect user privacy and confidentiality, we followed the relevant standard outlined by \citet{benton-etal-2017-ethical}. We removed all personally identifiable information, including usernames, metadata. Only meme images and their corresponding annotations were used for analysis.

This study received ethical approval from the University of Melbourne Human Research Ethics Committee (Project ID: 29894).

To minimize risks related to data distribution, we will release only the annotations and a retrieval script that enables approved researchers to re-download images directly from Reddit, in compliance with Reddit’s data policies \cite{reddit_policy_2024}. Access will be restricted to academic, non-commercial use.  The dataset (including splits used for analyses) is available at: \href{https://github.com/LiuliuChen/FigSIM}{https://github.com/LiuliuChen/FigSIM}.


\appendix

\newpage

\section{Pilot Rounds: Image Description and Annotation Note}
\label{sec:meme_explanation}

Two pilot rounds (50 memes each) were conducted to refine the annotation scheme. Four annotators (described in Section 3.3) independently annotated the pilot batches. Disagreements were resolved through regular adjudication meetings to ensure conceptual clarity and align interpretations. During these sessions, one annotator documented the adjudication discussions by writing short image descriptions (i.e., literal descriptions of the image and the meme template sources when identifiable) and summarizing the annotation notes for each meme. These notes were then reviewed and approved by a second annotator. 

\begin{figure}[htp]
    \centering
    \includegraphics[width=1\linewidth]{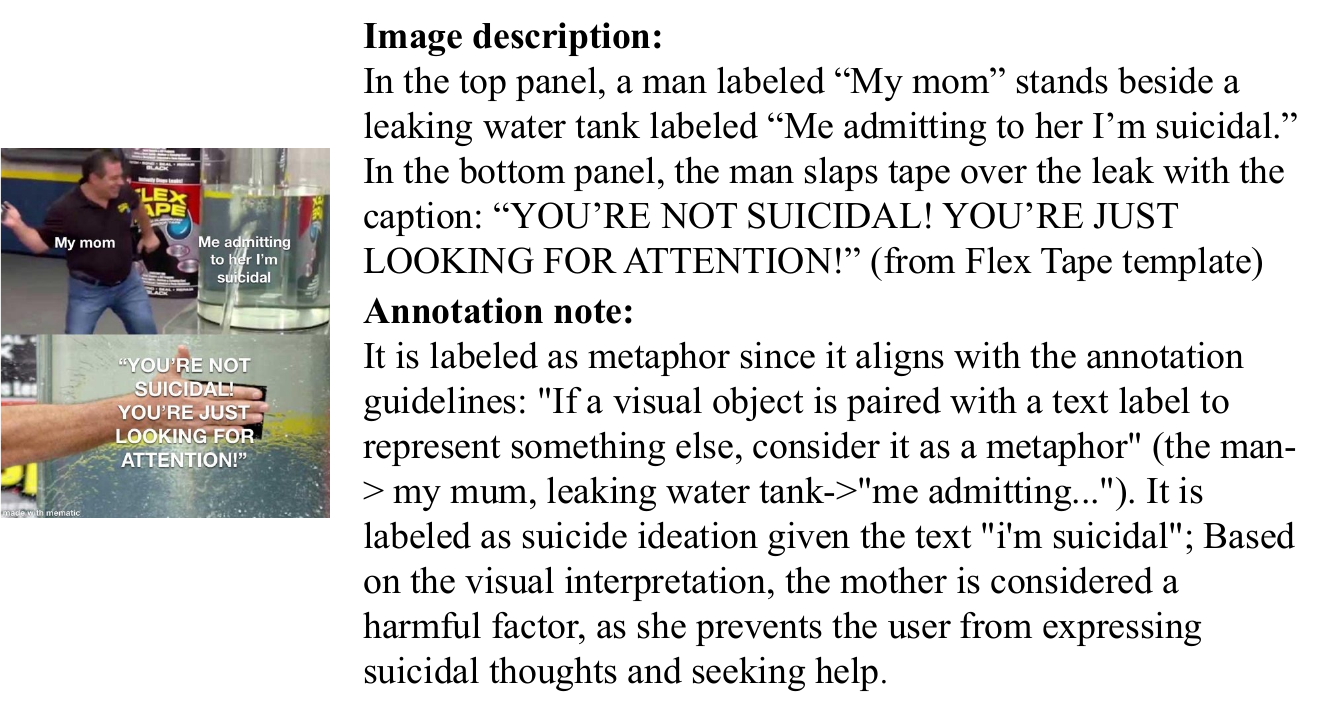}
    \caption{Examples of memes with image description and annotation note.}
    \label{fig:meme_text_exp}
\end{figure}

We emphasize that these explanations were created to clarify annotators’ interpretations during guideline development and calibration. They are not intended to be used as ground-truth reasoning, as more rigorous assessment would be required for that purpose. Figure \ref{fig:meme_text_exp} shows an example of the image descriptions and annotation notes.

\begin{figure*}[t]
    \centering
    \includegraphics[width=1\linewidth]{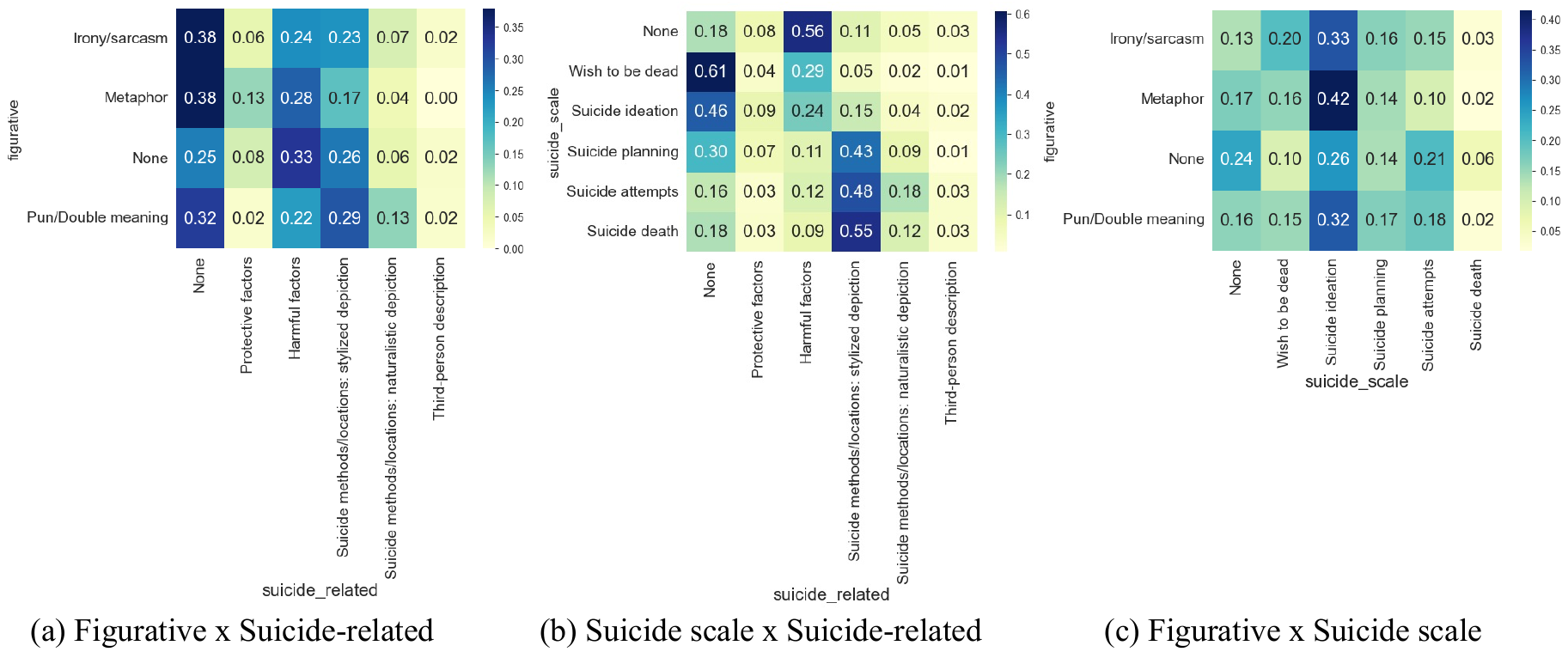}
    \caption{Cross-category heatmaps (normalized by row; darker colors indicate higher within-row proportions).}
    \label{fig:heatmap_cross_cat}
\end{figure*}

\section{Annotation Categories and Label Definitions}
\label{cat_def}

\subsection{Figurative Phenomenon}
\label{cat_def_fig}

The following are the definitions of labels in the Figurative Phenomenon category. Memes are labeled as \textbf{\textit{None} (F4)} if no figurative expressions apply.

\begin{itemize}[noitemsep, topsep=0pt]
    \item \textbf{Metaphor (F1):} A cross-domain mapping across conceptual domains (source to target). One domain of our experience (thoughts) can be understood in terms of another domain.
    \item \textbf{Pun/Double meaning (F2):} A form of expression that suggests multiple meanings through a term, similar-sounding words, or, in some cases, visual elements that create layered interpretations.
    \item \textbf{Irony/Sarcasm (F3):} Profess to hold beliefs or attitudes contrary to one’s underlying belief or attitude for humorous or emphatic effect.
\end{itemize}

\subsection{Suicide Severity Scale}
\label{cat_def_scale}

The following are the definitions of labels in the Suicide Severity Scale category. Memes are labeled as \textbf{\textit{None} (SS6)} if no severity applies.

\begin{itemize}[noitemsep, topsep=0pt]
        \item \textbf{Wish to be dead (SS1):} Thoughts about a wish to be dead or not alive anymore.
        \item \textbf{Suicide ideation (SS2):} Thoughts of wanting to end their life/die by suicide/kill oneself.
        \item \textbf{Suicide planning (SS3):} planning one's suicide such as thinking about or selecting methods, location or timing.
        \item \textbf{Suicide death (SS5):} Content that suggests a fatal outcome resulting from a suicide attempt.
    \end{itemize}

\subsection{Suicide-Related Content}
\label{cat_def_related}

The following are the definitions of labels in the Suicide-Related Content category. Memes are labeled as \textbf{\textit{None} (SR6)} if no suicide-related content is present.

\begin{itemize}[noitemsep, topsep=0pt]
        \item \textbf{Suicide methods/locations: stylized depiction (SR1):} Designed elements, symbolic patterns, or text description that could indicate a method/location of suicide.
        \item \textbf{Suicide methods/locations: naturalistic depiction (SR2):} Contain methods or locations where actual suicide could occur.
        \item \textbf{Protective factors (SR3):} Contain content that protects or prevents the user from suicidal thinking or acting on suicidal thoughts.
        \item \textbf{Harmful factors (SR4):} Contain content that reflects potential factors contributing to the emergence or worsening of suicidal thoughts.
        \item \textbf{Third-person description (SR5):} Mentions of suicide-related content that involve someone other than the author, which do not necessarily indicate the author’s suicidality.
    \end{itemize}

\section{Additional Dataset Statistics}
\label{sec: app_data_stat}
Figure \ref{fig:heatmap_cross_cat} shows heatmaps capturing pairwise relationships between suicide severity, figurative phenomenon, and suicide-related content. Figure \ref{fig:heatmap_cross_cat} (a) shows that figurative phenomenon co-occur more frequently with the labels \textit{None}, \textit{harmful factors}, and \textit{stylized method depictions} than with other suicide-related content labels. Figure \ref{fig:heatmap_cross_cat} (b) indicates that across all severity levels, a large proportion of memes are annotated with no suicide-related content, while stylized method depictions account for a higher proportion in higher-severity categories (e.g., \textit{suicide planning} and \textit{suicide attempts}) relative to other suicide-related labels. In Figure \ref{fig:heatmap_cross_cat} (c), all figurative categories show their highest proportions at the \textit{suicide ideation} level, while labels corresponding to \textit{suicide death} remain sparse.
Overall, these heatmaps reveal asymmetric relationships across annotation dimensions, reflecting overlap patterns within the dataset.

\section{Additional Inter-Annotators Agreement Analysis}
\label{sec: iaa}

In addition to the main IAA results reported in the dataset analysis section (Section~4.2), we provide supplementary agreement metrics for all annotation categories. Table~\ref{tab:overall_iaa_appendix} presents the observed agreement for the three single-label categories and the Jaccard similarity for the two multi-label categories. Observed agreement reflects the raw proportion of matching annotations, while Jaccard similarity captures set-level overlap for multi-label decisions.

\setlength{\tabcolsep}{1mm}
\begin{table}[!t]
\centering
{\fontsize{9pt}{9pt}\selectfont
\begin{tabularx}{\linewidth}{@{}l X X@{}}
\toprule
                                 & \textbf{Observed Agreement} & \textbf{Jaccard Similarity} \\ \midrule
\textbf{Context}                 & 0.92                        & \textbackslash{}            \\
\textbf{Modality}                & 0.98                        & \textbackslash{}            \\
\textbf{Suicide Severity Scale}  & 0.72                        & \textbackslash{}            \\
\textbf{Suicide-related Content} & \textbackslash{}            & 0.75                        \\
\textbf{Figurative Phenomenon}   & \textbackslash{}            & 0.78                        \\ \bottomrule
\end{tabularx}
}
\caption{Observed agreement and Jaccard similarity across annotation categories.}
\label{tab:overall_iaa_appendix}
\end{table}

To further examine consistency within multi-label categories, we additionally report label-wise Cohen’s~$\kappa$ scores in Tables~\ref{tab: iaa_figurative} and \ref{tab: iaa_suicide_content}. 

\setlength{\tabcolsep}{1mm}
\begin{table}[!t]
\centering
\fontsize{9pt}{9pt}\selectfont
\begin{tabularx}{\linewidth}{@{}l X X@{}}
\toprule
                              & \textbf{Cohen $\kappa$} & \textbf{Agreement} \\ \midrule
\textbf{Metaphor}             & 0.61                 & Substantial        \\
\textbf{Pun / Double meaning} & 0.57                 & Moderate           \\
\textbf{Irony/sarcasm}        & 0.52                 & Moderate           \\ \bottomrule
\end{tabularx}
\caption{Label-wise IAA: Figurative Phenomenon. Interpretation following \cite{landis1977measurement}.}
\label{tab: iaa_figurative}
\end{table}

\setlength{\tabcolsep}{1mm}
\begin{table}[!t]
\centering
\fontsize{9pt}{9pt}\selectfont
\begin{tabularx}{\linewidth}{@{}X c c@{}}
\toprule
                                                           & \textbf{Cohen $\kappa$} & \textbf{Agreement} \\ \midrule
\textbf{Suicide methods/locations: stylized depiction}     & 0.71                 & Substantial        \\
\textbf{Suicide methods/locations: naturalistic depiction} & 0.63                 & Substantial        \\
\textbf{Protective factors}                                & 0.62                 & Substantial        \\
\textbf{Harmful factors}                                   & 0.62                 & Substantial        \\
\textbf{Third-person description}                          & 0.59                 & Moderate           \\ \bottomrule
\end{tabularx}
\caption{Label-wise IAA: Suicide-related Content. Interpretation following \cite{landis1977measurement}.}
\label{tab: iaa_suicide_content}
\end{table}

\section{Implementation Details}
\label{sec: implementation_detail}

\subsection{Data Split}
We split the dataset into training, validation, and test sets using a stratified strategy to preserve label distributions across all annotation categories. The dataset is divided using a 60:20:20 ratio, resulting in 633 training memes, 208 validation memes, and 208 test memes. Table~\ref{tab:dataset_stats} reports detailed label-wise statistics for each split.

\setlength{\tabcolsep}{1mm}
\begin{table*}[!t]
\centering
{\fontsize{9pt}{9pt}\selectfont
\begin{tabular}{lcccccccccccccccccccccc}
\toprule
\multirow{2}{*}{\textbf{Split}} &
\multicolumn{4}{c}{\textbf{Figurative}} &
\multicolumn{6}{c}{\textbf{Suicide Severity Scale}} &
\multicolumn{6}{c}{\textbf{Suicide Related Content}} &
\multicolumn{3}{c}{\textbf{Modality}} &
\multicolumn{2}{c}{\textbf{Context}} &
\multirow{2}{*}{\textbf{Total}} \\
\cmidrule(lr){2-5}
\cmidrule(lr){6-11}
\cmidrule(lr){12-17}
\cmidrule(lr){18-20}
\cmidrule(lr){21-22}
 & F1 & F2 & F3 & F4
 & SS1 & SS2 & SS3 & SS4 & SS5 & SS6
 & SR1 & SR2 & SR3 & SR4 & SR5 & SR6
 & M1 & M2 & M3
 & C1 & C2
 & \\
\midrule
Train & 97 & 73 & 499 & 88 & 113 & 207 & 98 & 100 & 18 & 97 & 159 & 49 & 42 & 169 & 13 & 240 & 572 & 44 & 17 & 553 & 80 & 633 \\
Val   & 32 & 24 & 165 & 29 & 37  & 66  & 33 & 34  & 6  & 32 & 54  & 16 & 14 & 56  & 4  & 79  & 187 & 15 & 6  & 181 & 27 & 208 \\
Test  & 32 & 24 & 169 & 29 & 39  & 66  & 32 & 33  & 6  & 32 & 53  & 16 & 14 & 57  & 4  & 82  & 187 & 15 & 6  & 182 & 26 & 208 \\
\bottomrule
\end{tabular}
}
\caption{Dataset statistics across Figurative Phenomenon (F1–F4), Suicide Severity (SS1–SS6), Suicide-related Categories (SR1–SR6), Modality (M1–M3), and Context (C1–C2). Label codes (e.g., F1: Metaphor) and full definitions are provided in Appendix~\ref{cat_def}.}
\label{tab:dataset_stats}
\end{table*}

\subsection{Hyperparameters}
All supervised models are implemented in PyTorch. The hyperparameters used in our benchmark experiments are reported in Table~\ref{tab:hyperparameter}. For CLIP-based models, we use concatenation-based fusion of image and text representations. All models are trained using the same data splits to ensure fair comparison across configurations. Models are evaluated after each training epoch, and the checkpoint achieving the best macro F1 score on the validation set is saved. The checkpoint with the best validation macro F1 is then used for evaluation on the test set.

MLLMs are evaluated using zero-shot and few-shot prompting without task-specific fine-tuning. The decoding parameters are shown in Table~\ref{tab:llm_decoding}. Prompts are constructed using the category definitions and practical guidelines specified in the annotation scheme. Table~\ref{tab:fig_prompt} shows an example for figurative detection. For few-shot prompting, we randomly sample six instances for each task, ensuring that the examples cover all defined labels, and provide both images and their corresponding text labels as input to the models.

Experiments involving open-source and local models were run on NVIDIA A100 GPUs, while API-based LLMs were accessed via commercial endpoints.

\setlength{\tabcolsep}{1mm}
\begin{table}[!t]
\centering
\fontsize{9pt}{9pt}\selectfont
\begin{tabular}{@{}ll@{}}
\toprule
\textbf{Name} & \textbf{Value} \\ \midrule
Optimizer & AdamW \\
Learning rate & 2e-5 \\
Learning rate schedule & linear \\
Warmup ratio & 0.1 \\
Weight decay & 0.01 \\
Gradient norm clipping & 1.0 \\
Batch size & 16 \\
Epochs & [10, 20] \\ \bottomrule
\end{tabular}
\caption{Training hyperparameter setup.}
\label{tab:hyperparameter}
\end{table}

\setlength{\tabcolsep}{1mm}
\begin{table}[!t]
\centering
\fontsize{9pt}{9pt}\selectfont
\begin{tabular}{@{}ll@{}}
\toprule
\textbf{Parameter} & \textbf{Value} \\ \midrule
Decoding strategy & Deterministic (temperature=0) \\
Temperature & 0.0 \\
Top-p & 1.0 \\
Max output tokens & 128 \\
Stop sequences & \} (custom eos token) \\ \bottomrule
\end{tabular}
\caption{Decoding parameters for MLLM inference.}
\label{tab:llm_decoding}
\end{table}

\setlength{\tabcolsep}{1mm}
\begin{table*}[!t]
\centering
\fontsize{9pt}{9pt}\selectfont
\begin{tabular}{p{.9\linewidth}}
\toprule
\textbf{Figurative zero-shot prompt} \\ \midrule
You are annotating a SUICIDE-RELATED meme.\\ \\ TASK: Identify which figurative language the meme uses.\\ \\ Allowed labels (you may choose MULTIPLE):\\   - Metaphor\\   - Pun/Double meaning\\   - Irony/sarcasm\\ \\ DEFINITIONS:\\   - Metaphor: A cross-domain mapping across conceptual domains (source to target). One domain of our experience (thoughts) can be understood in terms of another domain.\\   - Pun/Double meaning: A form of expression that suggests multiple meanings through a term, similar-sounding words, or, in some cases, visual elements that create layered interpretations.\\   - Irony/sarcasm: Professing to hold beliefs or attitudes contrary to one’s underlying belief or attitude for humorous or emphatic effect.\\ \\ PRACTICAL GUIDELINES:\\   - Contrast effect in memes: Many memes use a contrast effect such as a happy image paired with a negative or dark caption, expectation vs. reality. This type of memes should be considered as Irony/sarcasm.\\   - Metaphor: If a visual object is paired with a text label (e.g., “me”, “life”) to represent something else, consider it as a metaphor. This style commonly appears in memes where meaning is constructed by assigning roles or abstract concepts to image elements. Exception: If the text label is simply adding context or clarification to the visual object, rather than representing something else conceptually, it should not be annotated as Metaphor.\\ \\ RESPONSE FORMAT:\\ Return ONLY a JSON object of the form:\\ \{"figurative": {[}\textless{}zero or more labels from the allowed list\textgreater{}{]}\}\\ \\ Rules:\\   - Use EXACT strings from the allowed labels (case-sensitive).\\   - If no figurative device is present, return an empty list: \{"figurative": {[}{]}\}\\   - Do NOT include any extra keys or text.\\ \midrule
\end{tabular}
\caption{Zero-shot prompt: figurative phenomenon detection}
\label{tab:fig_prompt}
\end{table*}

\section{Error Analysis}
\label{sec:app_err_analysis}

To better understand model performance, we conduct an error analysis on the best-performing models for each task: \texttt{Claude-sonnet-4-5 zero-shot} for figurative phenomenon detection, \texttt{Gemini~3~Pro few-shot} for suicide severity classification, and \texttt{Gemini~3~Pro zero-shot} for suicide-related content detection.

\begin{figure}[!t]
    \centering
    \includegraphics[width=\linewidth]{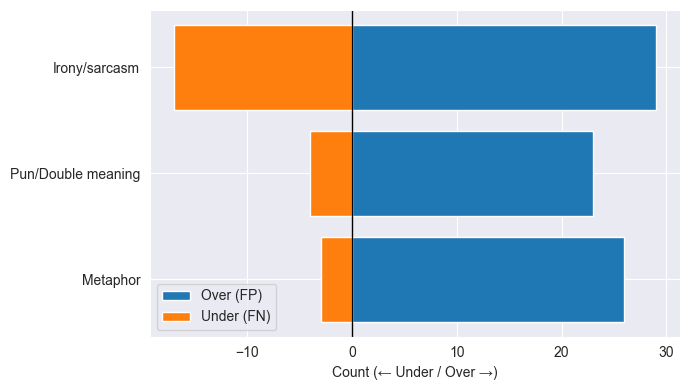}
    \caption{Over- and underprediction by figurative phenomenon.}
    \label{fig:fig_err}
\end{figure}

\begin{figure}[!t]
    \centering
    \includegraphics[width=\linewidth]{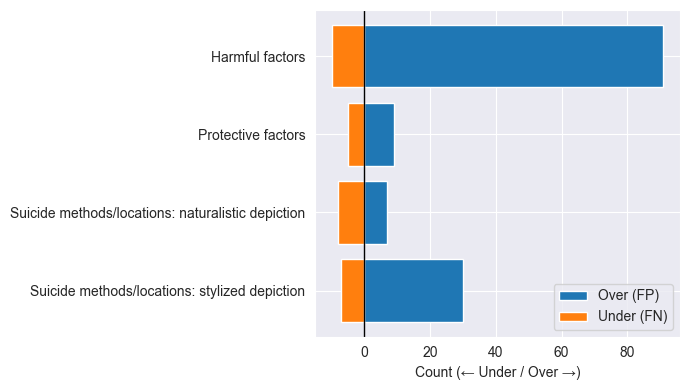}
    \caption{Over- and underprediction by suicide-related content.}
    \label{fig:related_err}
\end{figure}

\textbf{Label-wise Error Patterns}
Figures~\ref{fig:fig_err} and~\ref{fig:related_err} present label-wise over- and underprediction patterns for figurative phenomenon detection and suicide-related content detection, respectively. Here, over-prediction refers to false positive errors, where a label is predicted but not annotated, while underprediction refers to false negative errors, where an annotated label is not predicted by the model. Across both tasks, models tend to over-predict more salient or frequently occurring categories, such as \textit{irony/sarcasm} in figurative detection and \textit{harmful factors}, \textit{stylized depiction of suicide methods} in suicide-related content detection. Suicide severity prediction is analyzed separately in the main paper (Section~5.3), where over- and underprediction are defined with respect to the ordinal severity scale.

\textbf{Severity Underprediction by Figurative Subtype.}
Table~\ref{tab:severity_figurative} reports suicide severity underprediction rates conditioned on the presence of different figurative subtypes. Overall, severity is underpredicted more frequently when figurative language is present than when it is absent. Among figurative subtypes, metaphorical expressions exhibit the highest underprediction rate, followed by irony/sarcasm and puns/double meanings. We note that the number of metaphor and pun/double meaning samples in the test set is limited, and therefore subtype-level patterns should be interpreted descriptively rather than as definitive comparisons.

\begin{table}[t]
\centering
\small
\setlength{\tabcolsep}{6pt}
\begin{tabular}{lcc}
\toprule
\textbf{Figurative Property} & \textbf{Underprediction Rate} & \textbf{N} \\
\midrule
\textbf{Irony/Sarcasm} \\
\quad Absent & 0.429 & 7 \\
\quad Present & 0.604 & 48 \\
\midrule
\textbf{Metaphor} \\
\quad Absent & 0.551 & 49 \\
\quad Present & 0.833 & 6 \\
\midrule
\textbf{Pun/Double meaning} \\
\quad Absent & 0.571 & 49 \\
\quad Present & 0.667 & 6 \\
\bottomrule
\end{tabular}
\caption{Severity underprediction rates conditioned on figurative language properties. 
Underprediction is computed as the proportion of samples in which the predicted suicide severity is strictly lower than the annotated ground truth, treating severity labels as an ordered scale.}
\label{tab:severity_figurative}
\end{table}

\textbf{Context-required cases are challenging.} 
Due to dataset imbalance, most misclassified memes are \textit{no-context-required} and \textit{complementary} in absolute terms. However, within the misclassified test set, a substantial proportion of \textit{context-required} memes are misclassified across tasks (5/26 for suicide severity, 16/26 for figurative phenomenon, and 12/26 for suicide-related content detection). Similarly, while \textit{complementary} memes dominate the dataset, unimodal cases exhibit relatively higher error rates: 4/15 text-only memes are misclassified for suicide severity detection, 7/15 for figurative phenomenon detection, and 8/15 for suicide-related content detection. Additionally, 3/6 image-only memes are misclassified for suicide severity detection. These results are reported descriptively and suggest increased difficulty when contextual information or multimodal cues are limited, however, conclusions are constrained by the small number of such cases.

\begin{figure}[!t]
    \centering
    \begin{minipage}[t]{\linewidth}
    \subcaptionbox{}{
            \includegraphics[width=.95\linewidth]{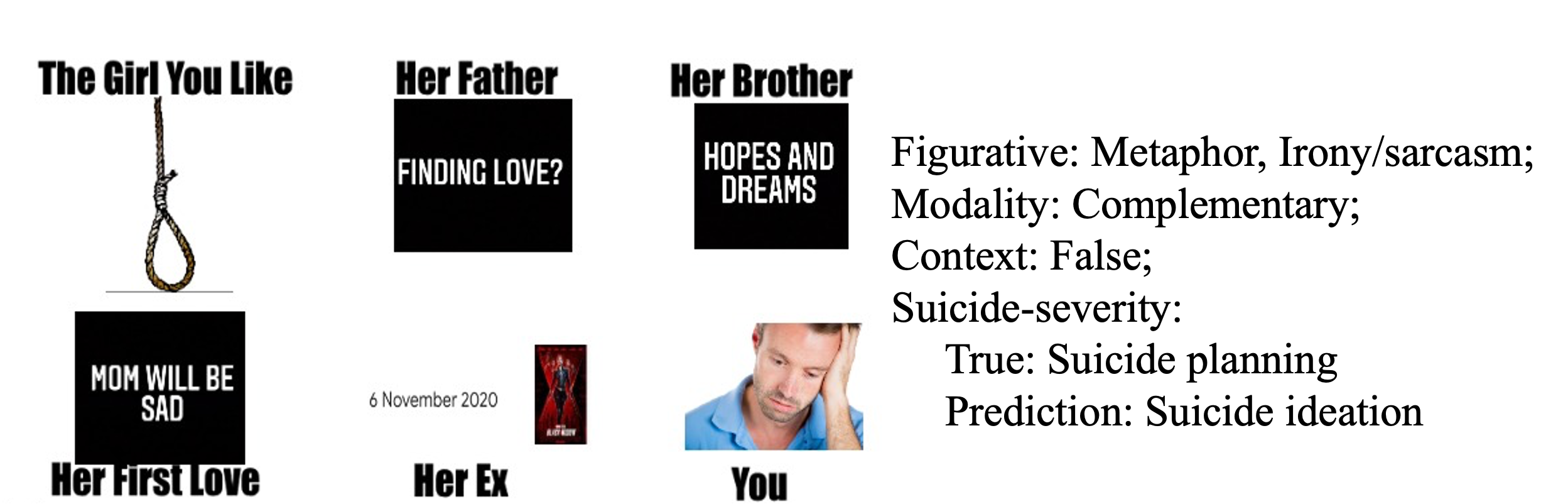}}  
    \end{minipage}
    \begin{minipage}[t]{\linewidth}
    \subcaptionbox{}{
            \includegraphics[width=.95\linewidth]{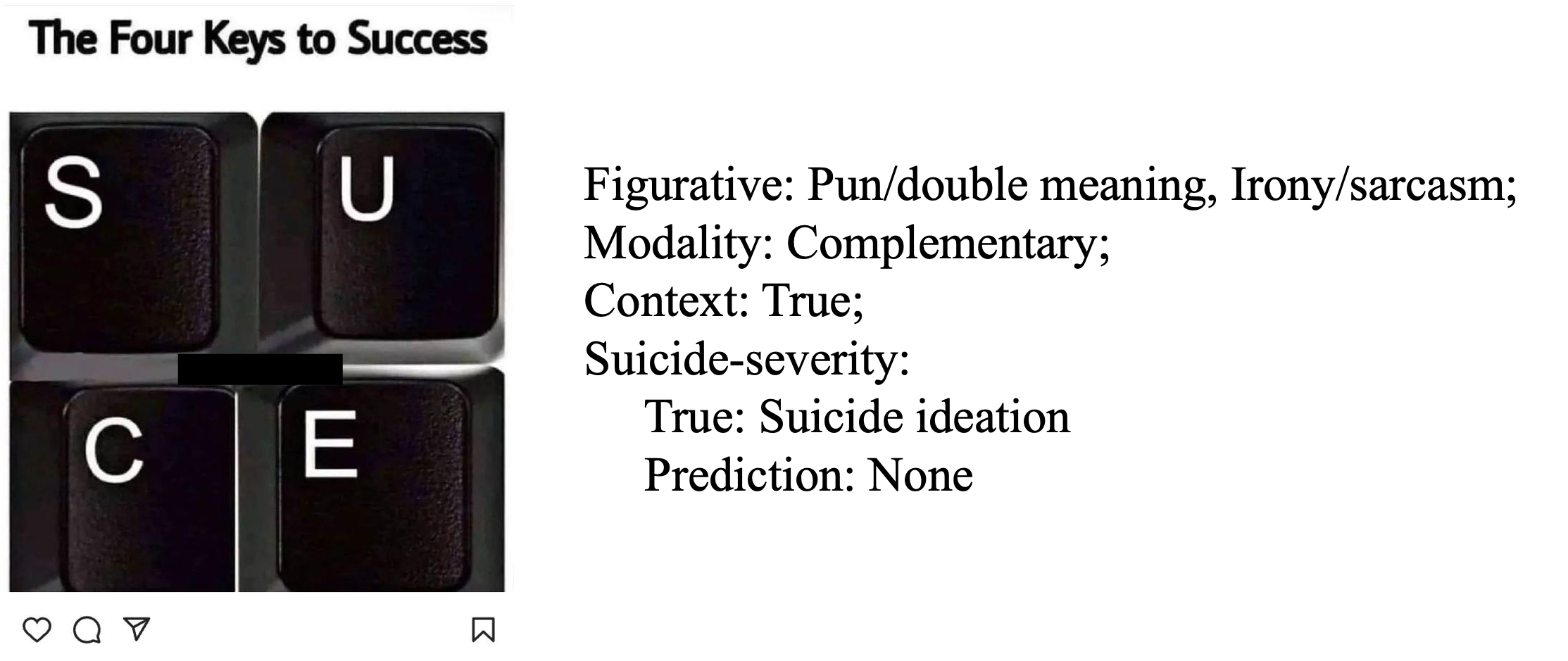}}       
    \end{minipage}
  \caption{Examples of misclassified cases}
  \label{fig:qual_err}
\end{figure}

\textbf{Qualitative Examples.}
Figure~\ref{fig:qual_err} shows examples of misclassified memes. In example (a), a suicide method is depicted metaphorically and ironically in a stylized format. Human annotators assigned the label \emph{Suicide planning}, as a specific method is mentioned, indicating the planning phase. Although the figurative labels are correctly identified, the model underpredicts severity, assigning \emph{Suicide ideation} instead of the annotated \emph{Suicide planning}. In example (b), suicidal ideation is expressed through a visual pun that relies on contextual interpretation (success vs.\ suicide), leading the model to predict \emph{None}. These examples illustrate how indirect, figurative, or context-dependent expressions can obscure severity cues.

\end{document}